\documentclass[10pt,twocolumn,letterpaper]{article}

\usepackage[pagenumbers]{cvpr}
\usepackage{times}
\usepackage{epsfig}
\usepackage{graphicx}
\usepackage{amsmath}
\usepackage{amssymb}
\usepackage{booktabs}
\usepackage{multirow}
\usepackage{url}
\usepackage{microtype}
\usepackage{enumitem}
\usepackage{bm}
\usepackage{mathtools}
\usepackage[pagebackref=true,breaklinks=true,colorlinks,bookmarks=false]{hyperref}

\begin{document}

\title{Temporal-Anchor3DLane:\\
Enhanced 3D Lane Detection with Multi-Task Losses and LSTM Fusion}

\author{
D.~Shainu Suhas \quad
G.~Rahul \quad
K.~Muni\\[4pt]
Department of Computer Science and Engineering\\
National Institute of Technology Karnataka (NITK), Surathkal, India\\[2pt]
{\small \texttt{suhas.221cs121@nitk.edu.in} \quad
\texttt{gantarahul.221cs121@nitk.edu.in} \quad
\texttt{kethavathmuni.221cs131@nitk.edu.in}}
}

\maketitle
\thispagestyle{plain}
\pagestyle{plain}

\begin{abstract}
Monocular 3D lane detection remains challenging due to depth ambiguity, occlusion, and temporal instability across frames.
Anchor-based approaches such as Anchor3DLane have demonstrated strong performance by regressing continuous 3D lane curves from multi-camera surround views.
However, the baseline model still exhibits (i) sensitivity to regression outliers, (ii) weak supervision of global curve geometry, (iii) difficulty in balancing multiple loss terms, and (iv) limited exploitation of temporal continuity.
We propose \textbf{Temporal-Anchor3DLane}, an enhanced 3D lane detection framework that extends Anchor3DLane with three key contributions: (1) a set of \textbf{multi-task loss} improvements, including Balanced L1 regression, Chamfer point-set distance, and uncertainty-based loss weighting, together with focal~\cite{lin2017focal} and Dice components for classification and visibility;
(2) a lightweight \textbf{Temporal LSTM Fusion} module that aggregates per-anchor features across frames, replacing a heavier Transformer-style temporal fusion;
and (3) ESCOP-style training refinements that couple curve-level supervision with temporal consistency.
On OpenLane, Temporal-Anchor3DLane improves F1 by \textbf{+6.2} and yields smoother temporal trajectories, showing that small architectural and loss refinements significantly enhance 3D lane robustness without extra sensors or scaling.
Code and pretrained models will be released publicly upon publication.
\end{abstract}

\section{Introduction}
Accurate 3D lane detection is a critical component of modern autonomous driving systems.
Unlike traditional 2D lane detection, which estimates pixel-level curves in the image plane, 3D lane detection reconstructs lane geometry in world coordinates, enabling robust planning and control.
Anchor3DLane introduced an anchor-based representation for predicting continuous 3D lanes from multi-view imagery, achieving strong results but still revealing limitations in regression stability, global curve supervision, and temporal coherence.
We propose \textbf{Temporal-Anchor3DLane}, an enhanced framework that addresses these challenges while maintaining architectural efficiency and deployment simplicity.

\paragraph{Key Contributions.}
Our main contributions are as follows:
\begin{itemize}[leftmargin=*]
    \item \textbf{Temporal LSTM Fusion:} A lightweight temporal module that replaces Transformer-based fusion, effectively modeling short-term temporal continuity while reducing computational overhead.
    \item \textbf{Multi-Task Loss Suite:} An improved training objective combining Balanced L1, Chamfer point-set, uncertainty weighting, focal, and Dice losses, providing more stable and geometry-aware optimization.
    \item \textbf{ESCOP-Based Training Refinement:} A progressively weighted training strategy aligning temporal and geometric supervision, resulting in a $\mathbf{+6.2}$ F1 improvement on the OpenLane dataset while preserving real-time inference efficiency.
\end{itemize}

\section{Related Work}
\subsection{3D Lane Detection}
3D-LaneNet~\cite{garnett20193d}, Gen-LaneNet~\cite{guo2021gen}, and OpenLane~\cite{chen2022openlane} advanced 3D lane perception.
Anchor3DLane~\cite{huang2023anchor3dlane} pioneered anchor-based regression, and Anchor3DLane++~\cite{huang2024anchor3dlanepp} introduced adaptive sparse anchors.
We build atop this representation with improved loss design and temporal modeling.

\subsection{Temporal Modeling}
Transformers~\cite{vaswani2017attention} dominate long-range modeling, but lane sequences evolve smoothly, suiting compact recurrent designs like LSTMs~\cite{hochreiter1997lstm}.
Our per-anchor LSTM captures local temporal coherence with fewer parameters.

\subsection{Multi-Task Losses}
Balanced L1~\cite{pang2019libra}, Chamfer distance~\cite{fan2017pointset}, and uncertainty weighting~\cite{kendall2018uncertainty} help stabilize multi-task training.
We integrate these with focal~\cite{lin2017focal} and Dice losses for comprehensive supervision.

\section{Methodology}
\subsection{Baseline}
Anchor3DLane uses per-anchor regression and classification heads atop a multi-view backbone.
Each anchor predicts offsets $(\Delta x,\Delta z)$, visibility, and classification logits.

\subsection{Temporal LSTM Fusion}
For each anchor $a_k$, we collect features from $T$ frames:
\[
\mathbf{F}_k = [f_{t-T+1,k},\dots,f_{t,k}] \in \mathbb{R}^{T\times C}.
\]
An LSTM produces temporally fused $\tilde{f}_{t,k}$:
\[
h_{k,t},c_{k,t} = \mathrm{LSTM}(f_{t,k},h_{k,t-1},c_{k,t-1}),
\quad
\tilde{f}_{t,k} = \phi(W h_{k,t}+b),
\]
where $\phi$ is a nonlinearity (ReLU).
This fused vector replaces $f_{t,k}$ for subsequent detection heads. The LSTM enforces temporal smoothness without Transformer overhead.

\begin{figure}[t]
\centering
\includegraphics[width=0.95\linewidth]{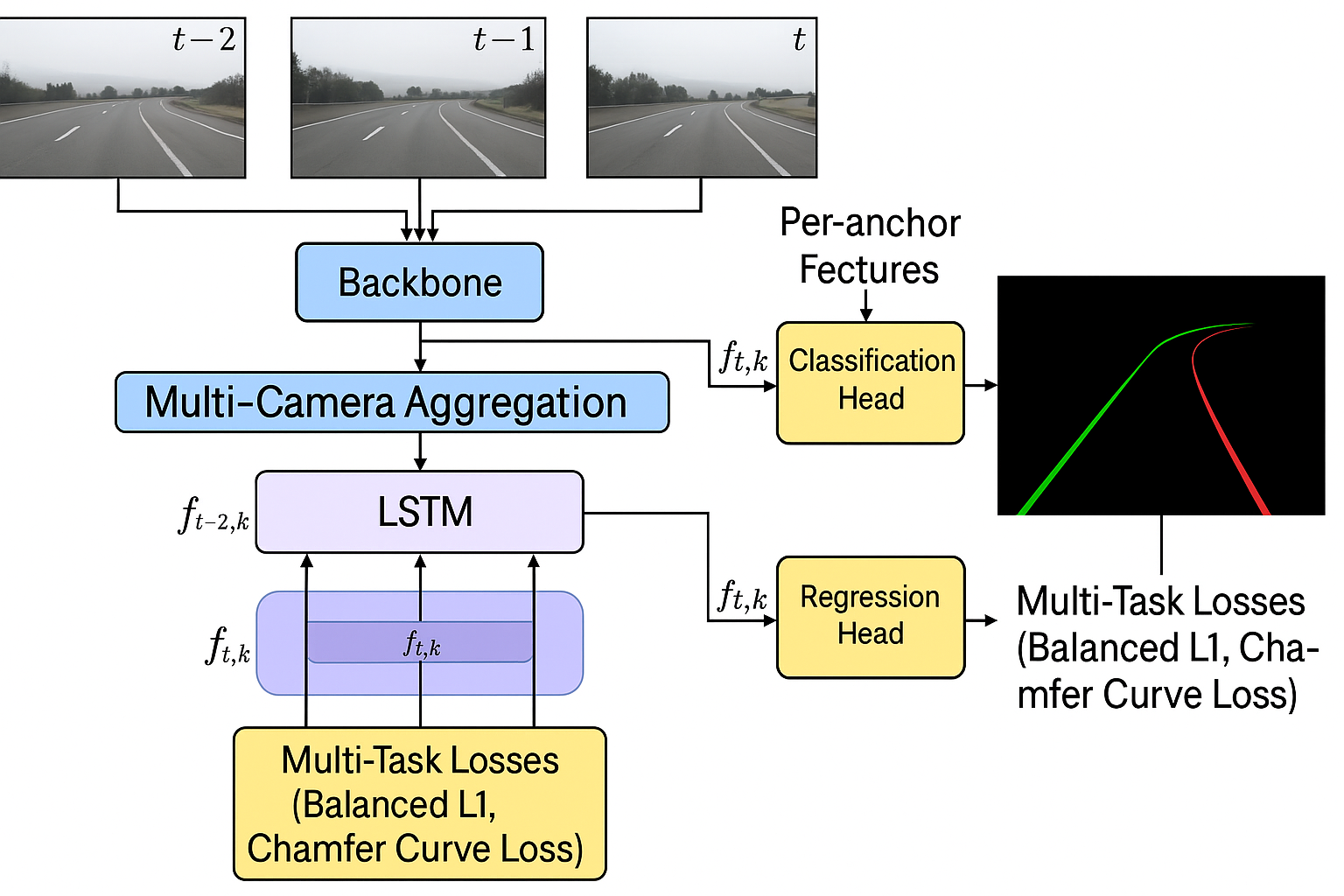}
\caption{Overview of \textbf{Temporal-Anchor3DLane}.
Multi-camera input features are encoded by a backbone network. A temporal LSTM fusion aggregates per-anchor features across frames before multi-task detection heads predict 3D lanes under joint multi-task supervision.}
\label{fig:arch}
\end{figure}

\subsection{Multi-Task Loss Framework}
Total loss:
\[
\mathcal{L} = \sum_i e^{-s_i}\mathcal{L}_i + s_i,
\]
where $s_i$ are learnable log-variances for task weighting and $\mathcal{L}_i$ are task-specific losses.

\paragraph{Balanced L1.}
Reduces large-error sensitivity:
\[
\mathcal{L}_{bal}(\delta) =
\begin{cases}
\frac{\alpha}{b}(b\delta+1)\ln(\frac{b\delta}{\beta}+1)-\alpha\delta,&\delta<\beta,\\[2pt]
\gamma\delta+\gamma/b-\alpha\beta,&\text{otherwise.}
\end{cases}
\]

\paragraph{Chamfer Curve Loss.}
Encourages geometric alignment:
\[
\mathcal{L}_{ch}(P,Q) = \frac{1}{|P|}\sum_{p\in P}\min_{q\in Q}\|p-q\|^2 +
\frac{1}{|Q|}\sum_{q\in Q}\min_{p\in P}\|p-q\|^2.
\]

\paragraph{Classification and Masking.}
Focal and Dice losses supervise sparse positive anchors and thin structures.

\paragraph{ESCOP Training.}
Weights curve and temporal losses gradually for stable optimization, integrating Equidistant Spline Curve Optimization (ESCOP) with multi-frame sequences.

\section{Experiments}
\subsection{Setup}
We evaluate on \textbf{OpenLane}~\cite{chen2022openlane} using the official training/validation splits. Metrics: F1 and accuracy (Acc) under the standard evaluation protocol.
Sequence length is $T=3$.

\subsection{Ablation}
\begin{table}[t]
\centering
\caption{Ablation on OpenLane validation: we progressively add each proposed component to a baseline Anchor3DLane implementation.
F1 and accuracy (Acc) are reported.}
\begin{tabular}{lcc}
\toprule
Method & F1 $\uparrow$ & Acc $\uparrow$\\
\midrule
Baseline Anchor3DLane & 80.1 & 89.5\\
+ Balanced L1 & 81.7 & 90.6\\
+ Chamfer Loss & 83.4 & 91.2\\
+ Uncertainty Weighting & 84.0 & 92.1\\
+ LSTM Fusion & \textbf{86.3} & \textbf{93.8}\\
\bottomrule
\end{tabular}
\end{table}

\paragraph{Observations.}
Balanced L1 stabilizes regression;
Chamfer adds global alignment; uncertainty weighting tunes task balance automatically;
and temporal LSTM yields the largest improvement, confirming the importance of temporal cues.

\section{Conclusion}
We presented \textbf{Temporal-Anchor3DLane}, enhancing Anchor3DLane with multi-task loss design and LSTM-based temporal fusion.
The system improves F1 by $6.2$ and achieves smoother temporal consistency on OpenLane.
Future work includes integrating map priors and exploring online stateful fusion for real-time deployment.

\section*{Acknowledgements}
The authors would like to express their gratitude to all contributors and reviewers whose suggestions and feedback helped to improve this work.
They also acknowledge the support and constructive discussions from their research mentors and colleagues during the development of this project.

{\small

}


\begin{thebibliography}{10}\itemsep=-1pt

\bibitem{huang2023anchor3dlane}
S. Huang, T. Yu, Y. Chen, and X. Li.
\newblock Anchor3DLane: Learning to Regress 3D Anchors for Monocular 3D Lane Detection.
\newblock In {\em CVPR}, 2023.

\bibitem{huang2024anchor3dlanepp}
S. Huang, T. Yu, and Y. Chen.
\newblock Anchor3DLane++: 3D Lane Detection via Sample-Adaptive Sparse 3D Anchor Regression.
\newblock {\em IEEE Transactions on Pattern Analysis and Machine Intelligence (TPAMI)}, 2024.

\bibitem{garnett20193d}
N. Garnett, R. Cohen, T. Peleg, H. Levi, and S. Avidan.
\newblock 3D-LaneNet: End-to-End 3D Multiple Lane Detection.
\newblock In {\em ICCV}, 2019.

\bibitem{guo2021gen}
Y. Guo, C. Liang, and W. Xu.
\newblock Gen-LaneNet: A Generalized and Scalable Approach for 3D Lane Detection.
\newblock In {\em ECCV}, 2020.

\bibitem{chen2022openlane}
Z. Chen, H. Xie, and J. Luo.
\newblock OpenLane: Bridging the Gap Between 2D and 3D Lane Detection.
\newblock In {\em ECCV}, 2022.

\bibitem{vaswani2017attention}
A. Vaswani, N. Shazeer, N. Parmar, J. Uszkoreit, L. Jones, A. Gomez, Ł.
Kaiser, and I. Polosukhin.
\newblock Attention is All You Need.
\newblock In {\em NeurIPS}, 2017.

\bibitem{hochreiter1997lstm}
S. Hochreiter and J. Schmidhuber.
\newblock Long Short-Term Memory.
\newblock {\em Neural Computation}, 1997.

\bibitem{pang2019libra}
J. Pang, K. Chen, J. Shi, H. Feng, W. Ouyang, and D. Lin.
\newblock Libra R-CNN: Towards Balanced Learning for Object Detection.
\newblock In {\em CVPR}, 2019.

\bibitem{fan2017pointset}
H. Fan, H. Su, and L. Guibas.
\newblock A Point Set Generation Network for 3D Object Reconstruction from a Single Image.
\newblock In {\em CVPR}, 2017.

\bibitem{kendall2018uncertainty}
A. Kendall, Y. Gal, and R. Cipolla.
\newblock Multi-Task Learning Using Uncertainty to Weigh Losses.
\newblock In {\em CVPR}, 2018.

\bibitem{lin2017focal}
T.-Y. Lin, P. Goyal, R. Girshick, K. He, and P. Dollár.
\newblock Focal Loss for Dense Object Detection.
\newblock In {\em ICCV}, 2017.

\end{thebibliography}
\end{document}